\documentclass[letterpaper, 10 pt, conference]{ieeeconf}  

\IEEEoverridecommandlockouts                      
\usepackage{xcolor, soul}
\usepackage{subcaption}
\usepackage{amsmath}
\usepackage{amsfonts}
\usepackage{amssymb}
\usepackage{bm}
\usepackage{multicol}
\usepackage{multirow}
\usepackage{optidef}
\usepackage{scalerel}
\usepackage{tikz}
\usetikzlibrary{svg.path}
\usepackage{colortbl}
\usepackage{color}
\definecolor{celadon}{rgb}{0.67, 0.88, 0.69}
\usepackage{algorithm}
\usepackage{algpseudocode}
\usepackage{lipsum}
\usepackage{titlesec}
\usepackage{hyperref}

\definecolor{orcidlogocol}{HTML}{A6CE39}
\tikzset{
  orcidlogo/.pic={
    \fill[orcidlogocol] svg{M256,128c0,70.7-57.3,128-128,128C57.3,256,0,198.7,0,128C0,57.3,57.3,0,128,0C198.7,0,256,57.3,256,128z};
    \fill[white] svg{M86.3,186.2H70.9V79.1h15.4v48.4V186.2z}
                 svg{M108.9,79.1h41.6c39.6,0,57,28.3,57,53.6c0,27.5-21.5,53.6-56.8,53.6h-41.8V79.1z M124.3,172.4h24.5c34.9,0,42.9-26.5,42.9-39.7c0-21.5-13.7-39.7-43.7-39.7h-23.7V172.4z}
                 svg{M88.7,56.8c0,5.5-4.5,10.1-10.1,10.1c-5.6,0-10.1-4.6-10.1-10.1c0-5.6,4.5-10.1,10.1-10.1C84.2,46.7,88.7,51.3,88.7,56.8z};
  }
}

\newcommand\orcidiconKLW[1]{\href{https://orcid.org/0000-0002-1938-4222}{\mbox{\scalerel*{
\begin{tikzpicture}[yscale=-1,transform shape]
\pic{orcidlogo};
\end{tikzpicture}
}{|}}}}
\newcommand\orcidiconAAS[1]{\href{https://orcid.org/0000-0002-6140-619X}{\mbox{\scalerel*{
\begin{tikzpicture}[yscale=-1,transform shape]
\pic{orcidlogo};
\end{tikzpicture}
}{|}}}}
\newcommand\orcidiconAK[1]{\href{https://orcid.org/0000-0002-3494-0469}{\mbox{\scalerel*{
\begin{tikzpicture}[yscale=-1,transform shape]
\pic{orcidlogo};
\end{tikzpicture}
}{|}}}}
\newcommand\orcidiconFGS[1]{\href{https://orcid.org/0000-0002-5090-9007}{\mbox{\scalerel*{
\begin{tikzpicture}[yscale=-1,transform shape]
\pic{orcidlogo};
\end{tikzpicture}
}{|}}}}
\newcommand\orcidiconRG[1]{\href{https://orcid.org/0000-0001-9701-879X}{\mbox{\scalerel*{
\begin{tikzpicture}[yscale=-1,transform shape]
\pic{orcidlogo};
\end{tikzpicture}
}{|}}}}
\newcommand\orcidiconSA[1]{\href{https://orcid.org/0000-0002-1736-5829}{\mbox{\scalerel*{
\begin{tikzpicture}[yscale=-1,transform shape]
\pic{orcidlogo};
\end{tikzpicture}
}{|}}}}
\newcommand\orcidiconYC[1]{\href{https://orcid.org/0000-0001-5767-7029}{\mbox{\scalerel*{
\begin{tikzpicture}[yscale=-1,transform shape]
\pic{orcidlogo};
\end{tikzpicture}
}{|}}}}

\title{\LARGE \bf
Exploiting Disturbance Preview for Predictive Trajectory Tracking of Underwater Vehicles in Wave Dominated Environments
}
\title{\LARGE \bf
Disturbance Preview for Nonlinear Model Predictive Trajectory Tracking of Underwater Vehicles in Wave Dominated Environments}

\author{Kyle L. Walker$^{1}$\orcidiconKLW{0000-0002-1938-4222}, \IEEEmembership{Student Member, IEEE,} 
        and Francesco Giorgio-Serchi$^{1}$\orcidiconFGS{0000-0002-5090-9007}
\thanks{Manuscript received XXXX; revised XXXX.}
\thanks{This work was supported by the EPSRC under grant No. EP/R026173/1 and grant No. EP/R513209/1.}
\thanks{$^{1}$Kyle L. Walker and Francesco Giorgio-Serchi are with the Institute for Integrated Micro and Nano Systems, School of Engineering, University of Edinburgh, Edinburgh,
U.K. (Correspondence: F.Giorgio-Serchi@ed.ac.uk).}
\thanks{Digital Object Identifier (DOI): see top of this page.}
}

\begin{document}

\maketitle
\thispagestyle{empty}
\pagestyle{empty}

\begin{abstract}

Operating in the near-vicinity of marine energy devices poses significant challenges to the control of underwater vehicles, predominantly due to the presence of large magnitude wave disturbances causing hazardous state perturbations. Approaches to tackle this problem have varied, but one promising solution is to adopt predictive control methods. Given the predictable nature of ocean waves, the potential exists to incorporate disturbance estimations directly within the plant model; this requires inclusion of a wave predictor to provide online preview information. To this end, this paper presents a Nonlinear Model Predictive Controller with an integrated Deterministic Sea Wave Predictor for trajectory tracking of underwater vehicles. State information is obtained through an Extended Kalman Filter, forming a complete closed-loop strategy and facilitating online wave load estimations. The strategy is compared to a similar feed-forward disturbance mitigation scheme, showing mean performance improvements of $\mathbf{51\%}$ in positional error and $\mathbf{44.5\%}$ in attitude error. The preliminary results presented here provide strong evidence of the proposed method's high potential to effectively mitigate disturbances, facilitating accurate tracking performance even in the presence of high wave loading.

\end{abstract}

\section{Introduction}

Increasing the level of autonomy is becoming a key area of interest in the marine sector \cite{Shukla2016}, motivated by the growing need to perform frequent maintenance procedures on devices and plants located in areas susceptible to adverse weather conditions \cite{Khalid2022}. For example, offshore renewable devices are required to be functional in highly energetic environments to operate effectively, however this can be problematic when deploying robotic systems for inspection and maintenance as they can be heavily influenced by fluid-body interactions. This raises concerns over the ability to avoid collisions when operating at close-proximity with submerged structures. In these environments, ocean waves can display wave heights several times the vehicle length, with their effects propagating through the water column \cite{FossenBook}. This poses significant threats to system safety at lower depths when inspection and maintenance tasks are necessary.


To tackle this problem, typical approaches have aimed to formulate robust control schemes which can mitigate general state perturbations effectively, for instance using sliding mode \cite{Garcia-Valdovinos2014, VonBenzon2021} or adaptive control \cite{Barbalata2015}. These methods have been shown to function well for generalised disturbances, however control performance may degrade under large magnitude, highly unsteady hydrodynamic loads. Other approaches have suggested performing load estimation \emph{in-situ} \cite{Riedel1998b}, but limitations relating to the reactive nature of the underlying control remain. Alternatively, predictive control strategies can reduce this limitation; under this scope, varying forms of Model Predictive Control (MPC) have been proposed \cite{Heshmati-Alamdari2021, Shen2015, Shen2017}. For underwater navigation tasks, the optimal nature of the MPC structure is especially effective \cite{Wei2023}, but the contribution from hydrodynamic disturbances is often neglected with operation in calmer waters usually considered \cite{Caldwell2010, Wallen2019} or focus is placed on disturbances of a quasi-constant nature such as ocean currents \cite{Wei2021}. In contrast, when wave disturbances are explicitly incorporated in the control, the proposition typically lacks a realistic estimation method, assuming the future loading is given \cite{Fernandez2017}. As with other control methods, larger magnitude and/or unknown disturbances can potentially lead to instabilities \cite{Fang2022}; however, ocean waves exhibit a degree of predictability which facilitates their analytical representation. Therefore, incorporation of reconstructed disturbances in the form of preview information offers a viable solution to improving control robustness, stability and real-world applicability \cite{Monasterios2018}.

Obtaining preview information in this instance requires a wave predictor to produce short term forecasts continuously, exploiting these within the MPC optimisation. Throughout the literature, typical temporal forecasting methods have included the use of auto-regressive models \cite{Fusco2009, Ge2016, PenaSanchez2020} or deterministic methods \cite{Belmont2006, Belmont2014, Al-Ani2020}, the latter shown to be effective at predicting the wave elevation over short-term horizons by explicitly considering low-order models. These deterministic approaches exploit measurements related to the wave profile at a particular point in space or time, forming future predictions at an alternative point. As predictions are based purely on in-situ measurements, the process remains agnostic to the sea state.  Alternatives have investigated learning-based methods \cite{Malekmohamadi2011, Pirhooshyaran2020}, however these typically relate to statistical predictions over longer time-periods. The majority of techniques which involve a deterministic wave prediction have been tailored towards maximising power output from wave energy converters \cite{Ling2015, Li2014}, nevertheless this same approach lends itself to dynamic positioning of underwater vehicles. Preview of disturbances must be obtainable in a quick and \emph{ad-hoc} manner, therefore load estimation models based on spectral analysis are ideal \cite{WalkerRAL, WalkerICRA}. Given the established nature of state estimation techniques \cite{Long2021}, this final element can be incorporated to produce a highly effective control framework for mitigating wave-induced disturbances to underwater vehicles.

\begin{figure}[t!]
    \centering
    \includegraphics[width=0.48\textwidth]{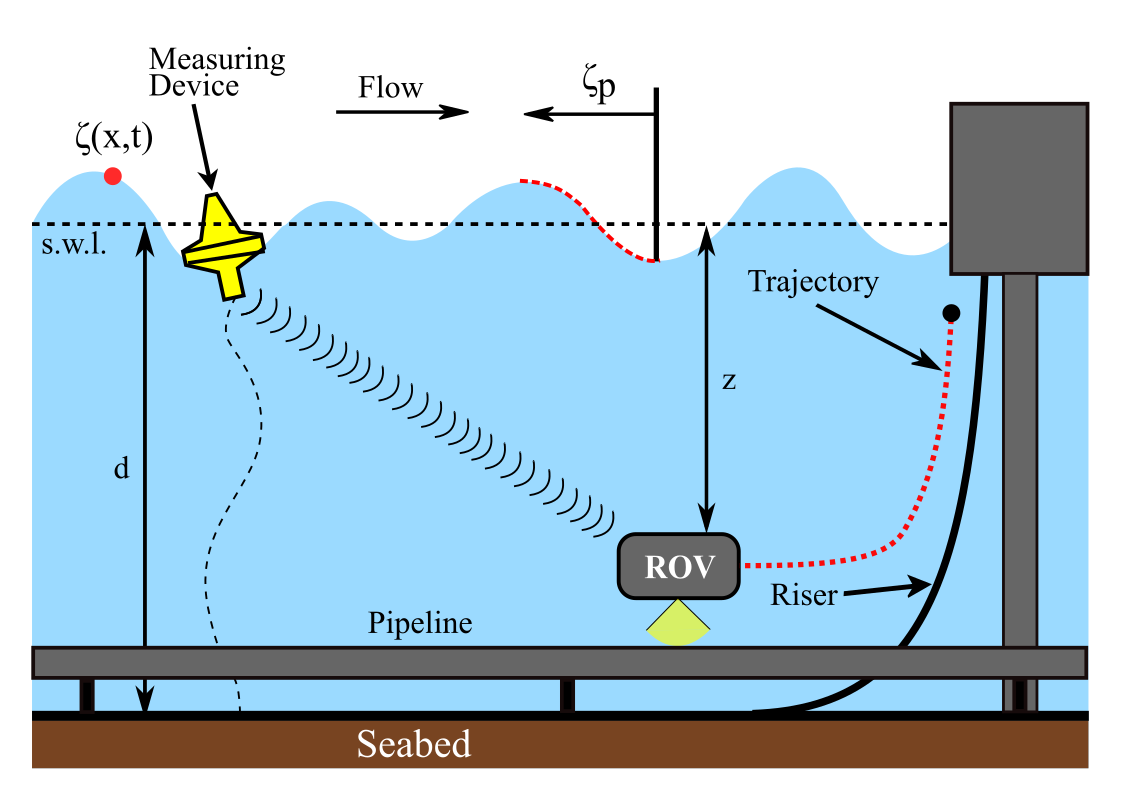}
    \caption{Schematic representation of the proposed control strategy; data is accumulated by the measuring device, subsequently passing disturbance preview information to the vehicle controller to mitigate state perturbations.}
    \label{art_schematic}
\end{figure}

In this paper, an architecture for trajectory tracking of an underwater vehicle subjected to wave disturbances is proposed. A Nonlinear Model Predictive Controller (NMPC) featuring a Deterministic Sea Wave Prediction (DSWP) algorithm is implemented, providing real-time disturbance preview information to the controller. The proposed control framework encompasses the following: (1) wave measurements are monitored at a fixed-location distant from the vehicle operating point ($50m$); (2) subsequently, a wave predictor propagates spectral information to the vehicle's location for predicting impending wave loads on the vehicle (see Fig. \ref{art_schematic}); and (3) the NMPC exploits these predictions in conjunction with the vehicle dynamic model to produce an optimised set of disturbance mitigating control actions. An Extended Kalman Filter (EKF) provides the necessary state information  (namely the positional data of the ROV) required by the wave predictor and vehicle controller, forming a closed-loop framework. The proposed control is directly compared to both a typical feedback based controller and a disturbance compensating feed-forward controller, the latter also exploiting the DSWP estimations. The NMPC outperforms both for a set of three unique sea states, presenting evidence that exploiting preview information within a NMPC has the potential to reduce state perturbations, offering a solution to the subsea vehicle control problem in harsh environments.  

\section{Modelling}

In our analysis, we consider the scenario to be a 3 Degree-of-Freedom (DoF) system concerning motion in the surge, heave and pitch. This stems from modelling the sea surface as a uni-directional wave to present an initial evaluation of controller performance combined with a planar prediction model, analogous to the scenario depicted in Fig. \ref{art_schematic}.


\subsection{Rigid-Body Vehicle Dynamics}

\begin{figure}[t!]
    \centering
    \includegraphics[width=0.45\textwidth]{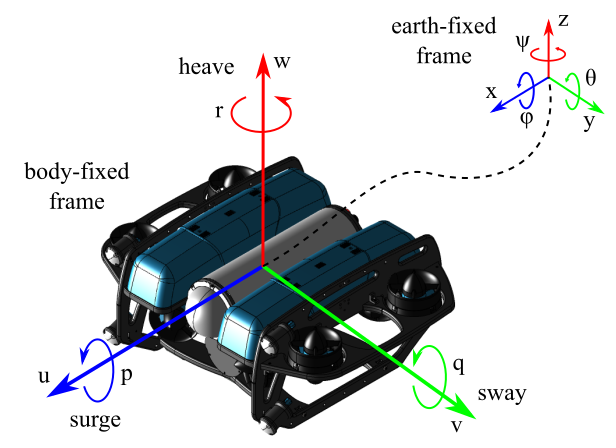}
    \caption{The BlueROV2 Heavy, displaying notations in conjunction with the earth-fixed and body-fixed co-ordinate frames.}
    \label{frames}
\end{figure}

The relationship between the earth-fixed frame and the body-fixed frame shown in Fig. \ref{frames} is defined by
\begin{equation}
    \dot{\boldsymbol{\eta}} = \mathbf{J(\boldsymbol{\eta})}\bm{\nu}
\end{equation}
\noindent where $\boldsymbol{\eta} = [x, z, \theta]^{T}$ is a state vector describing the position and orientation of the vehicle, $\bm{\nu} = [u, w, q]^{T}$ is a state vector of linear and angular velocities and $\mathbf{J}\in \mathbb{R}^{3\times 3}$ is a transformation matrix relating the two frames. 

Through this kinematic description, the nonlinear dynamics of the vehicle can be defined according to rigid-body theory, resulting in:
\begin{equation} \label{dynamic_equation}
    \mathbf{M\dot{\boldsymbol{\nu}}} + \mathbf{C(\boldsymbol{\nu})\boldsymbol{\nu}} + \mathbf{D(\boldsymbol{\nu})\boldsymbol{\nu}} + \mathbf{g(\boldsymbol{\eta})} = \boldsymbol{\tau} + \boldsymbol{\tau}_{E}
\end{equation}
\noindent where $\mathbf{M} = \mathbf{M}_{RB} + \mathbf{M}_{A} \in \mathbb{R}^{3\times3}$ is an inertia matrix, $\mathbf{C(\boldsymbol{\nu})}  = \mathbf{C_{RB}(\boldsymbol{\nu})} + \mathbf{C_{A}(\boldsymbol{\nu})} \in \mathbb{R}^{3\times3}$ is a matrix of Coriolis and centripetal terms, $\mathbf{D(\boldsymbol{\nu})} \in \mathbb{R}^{3\times3}$ is a hydrodynamic damping matrix, $\mathbf{g(\boldsymbol{\eta})} \in \mathbb{R}^{3}$ is a vector of hydrostatic restoring forces, $\boldsymbol{\tau}\in \mathbb{R}^{3}$ is a vector of external forces and moments (e.g. thrust) and $\boldsymbol{\tau}_{E}\in \mathbb{R}^{3}$ is a vector of environmental disturbances. In the above, subscripts $_{RB}$ and $_A$ relate to contributions from rigid body and added inertial effects. Throughout this work, only the effects of surface waves are considered to contribute to $\boldsymbol{\tau_{E}}$, as the vehicle is assumed to be operating in the upper portion of the water column where these effects are more prevalent.

\subsection{Wave-induced Hydrodynamic Disturbances}
 
Adopting the principle of linear superposition, the sea state can be described by a spectrum of $N$ monochromatic components, each with a unique wave height, $H$, period, $T$, and phase offset, $\epsilon$, producing a 2nd-order description of the surface elevation at a point $(x,t)$ in space and time \cite{FossenBook}:
\begin{align}
\zeta(x, t) = & \sum_{i=0}^{N} \frac{H_{i}}{2}cos(\kappa_{i}x - \omega_{i} t + \epsilon_{i}) \nonumber \\
  & + \sum_{i=1}^{N} \frac{1}{2}\kappa_{i}\left(\frac{H_{i}}{2}\right)^{2} cos2(\kappa_{i}x - \omega_{i} t + \epsilon_{i}))
\label{WaveEquation}
\end{align}
where $\kappa$ and $\omega$ represent the wave number and the angular frequency respectively. Solving for these parameters requires consideration of the dispersion relation:
\begin{equation}
    \omega_{i} = \sqrt{g\kappa_{i} \tanh{\kappa_{i}d}}
\end{equation}
where $g$ and $d$ are the gravitational constant and seabed depth. Analogously, the particle velocities in the global frame can be deduced for the surge and heave components \cite{McCormickBook}, such that:
\begin{align}
	\label{ParticleVelocityX}
	& u_{p}(x, z, t) = \sum_{i=1}^{N} \frac{g H_{i}}{2c}\frac{\cosh \kappa_{i}(z+d)}{\cosh \kappa_{i}d} \cos(\kappa_{i}x - \omega_{i} t + \epsilon_{i}) \nonumber &\\
         & + \frac{3}{16}c\kappa_{i}^2H_{i}^2 \frac{\cosh[2\kappa_{i}(z+d)]}{\sinh^{4} \kappa_{i}d} \cos[2(\kappa_{i}x - \omega_{i} t + \epsilon_{i})]
\end{align}	
\begin{align}
	\label{ParticleVelocityZ}
	& w_{p}(x, z, t) = \sum_{i=1}^{N} \frac{g H_{i}}{2c}\frac{\sinh \kappa_{i}(z+d)}{\cosh \kappa_{i}d} \sin(\kappa_{i}x - \omega_{i} t + \epsilon_{i})\nonumber &\\
        & + \frac{3}{16}c\kappa_{i}^2H_{i}^2 \frac{\sinh[2\kappa_{i}(z+d)]}{\sinh^{4} \kappa_{i}d} \sin[2(\kappa_{i}x - \omega_{i} t + \epsilon_{i})]
\end{align}

\noindent where $z$ and $c$ are the evaluated depth and wave celerity respectively. Given the magnitude of the fluid velocity components, a low-order model can be deployed to estimate the wave-induced hydrodynamic disturbances by considering the added inertia and hydrodynamic drag loading. Defining $\boldsymbol{\nu}_{p} = \left[ \nu_{p,x}, \nu_{p,z} \right]^{T} = \mathbf{R}_{y}(\theta)^{T}\left[ u_{p}, w_{p} \right]^{T}$ (where $\mathbf{R}_{y}(\theta)$ is a rotation matrix), it follows that:
\begin{equation} \label{wave_forces}
    \boldsymbol{\tau}_{E} = \begin{bmatrix}
        X_{E} \\ Z_{E} \\ M_{E}
    \end{bmatrix} = 
    \begin{bmatrix}
        X_{\dot{u}}\dot{\nu}_{p,x} + \lbrace X_{u} + X_{u|u|}|\nu_{p,x}| \rbrace \nu_{p,x} \\ Z_{\dot{w}}\dot{\nu}_{p,z} + \lbrace Z_{w} + Z_{w|w|}|\nu_{p,z}| \rbrace \nu_{p,z} \\ 
        \int_{-L/2}^{L/2} Z_{E}(x',z',t)x' dx
    \end{bmatrix}
\end{equation}
where $L$ is the vehicle body length and $(x',z')$ refers to the point of evaluation in the vehicle's local frame. This methodology was validated experimentally \cite{WalkerRAL, WalkerICRA, Gabl2020, Gabl2021} and is motivated by the desire for computationally inexpensive estimations of the wave-induced hydrodynamic loads, suitable for inclusion within a predictive control architecture where fast calculation is critical.


\section{Nonlinear Model Predictive Control with Preview}

\begin{figure}[t!]
    \centering
    \includegraphics[width=0.48\textwidth]{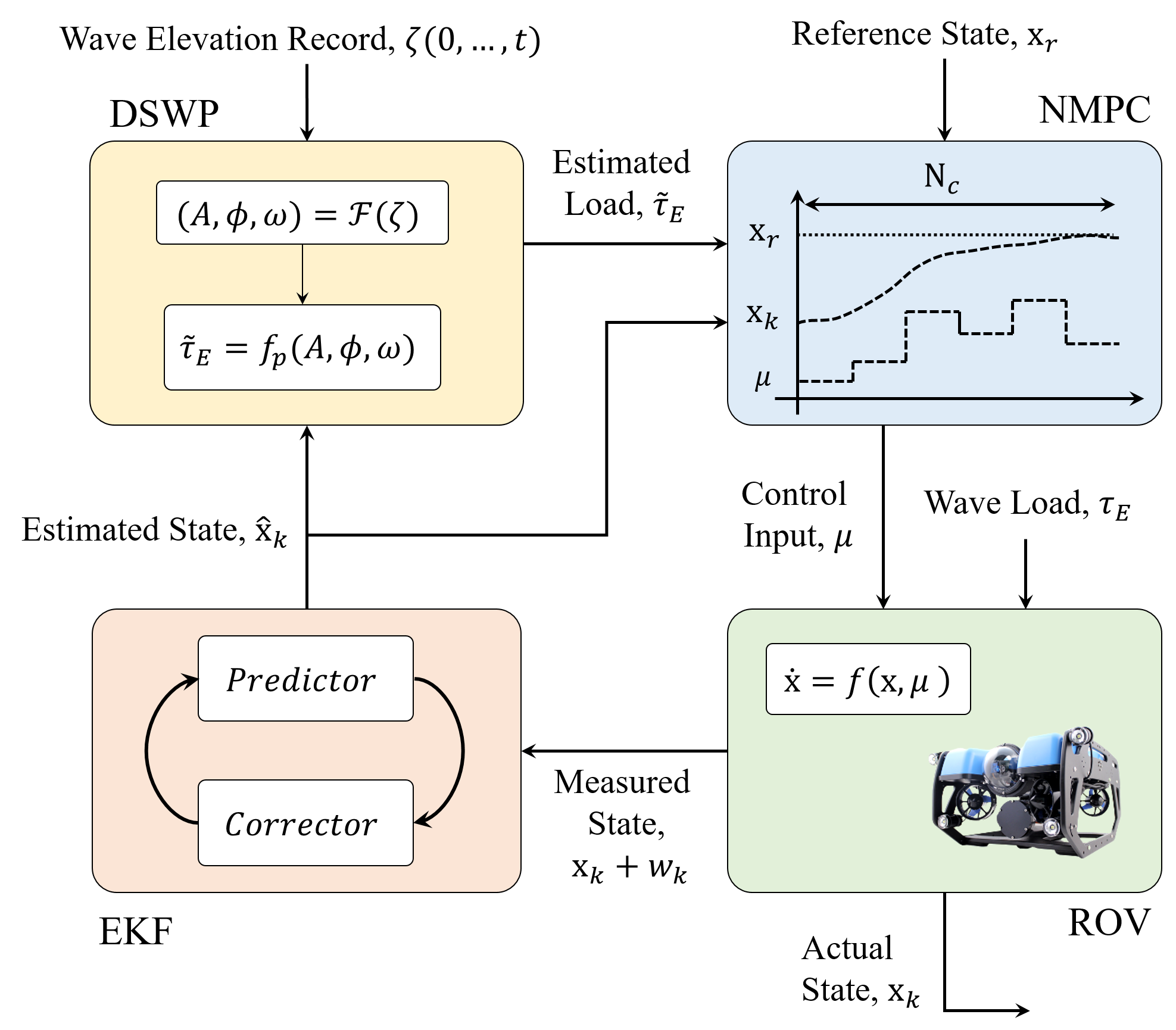}
    \caption{Block diagram representation of the proposed predictive control scheme with disturbance preview information incorporated.}
    \label{block_diagram}
\end{figure}

As the proposed control method in this work is reliant on knowing the vehicle position and measurement location to produce preview information of disturbances, the controller was formulated in conjunction with an EKF as a state estimator for the ROV, assuming some knowledge of the vehicle state could be obtained through sensor measurements. Wave buoy's are typically GPS triangulated to a high degree of accuracy, thus the accuracy of the state estimation is predominantly dependent on sensor quality which can be maximised by choosing high performance components. Deduction of the full vehicle state facilitates the remaining stage of the closed-loop framework, providing the necessary information for both the disturbance predictor and NMPC to function as one. Fig. \ref{block_diagram} outlines a block-diagram representation of the proposed framework.




\subsection{Disturbance Preview Predictor}
\label{wave_predictor}
Naturally, ocean waves have a prescribed degree of predictability associated with their occurrence; this can be exploited to provide the NMPC with a preview of oncoming disturbances along the controller prediction horizon to improve control. To achieve this, we employ a fixed-point Deterministic Sea Wave Prediction (DSWP) algorithm \cite{Belmont2006}, which monitors the sea surface elevation for a period of time $T_{M}$ at a specified distance from the vehicle location $x_{\scriptscriptstyle P}$. This is used to form short term predictions of the future wave elevations and subsequently an estimation of the wave-induced disturbances. When deploying DSWP, the sea surface is assumed to be statistically stationary over the time duration of the prediction process, which for short-term horizon predictions holds \cite{Li2012}. This allows exploitation of the sea state spectral properties for the purposes of forming predictions with a reliable degree of accuracy.

\begin{figure}[t!]
    \centering
    \includegraphics[width=0.48\textwidth]{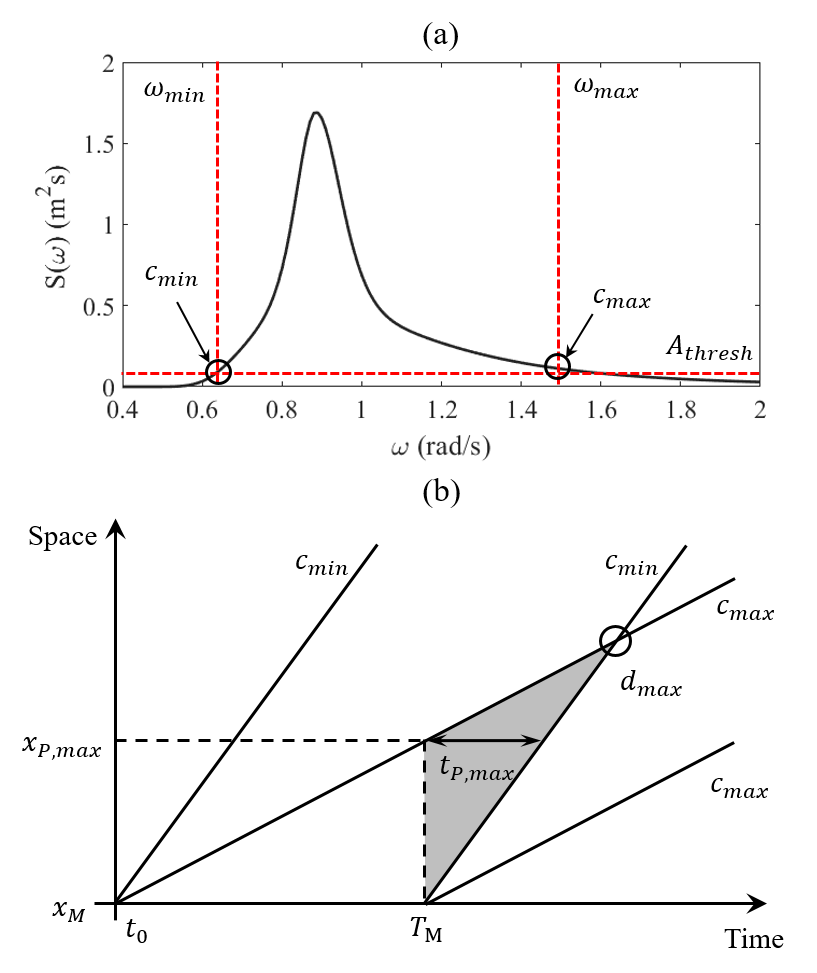}
    \caption{(a) The wave frequency spectrum with frequency and amplitude thresholds applied to bound the components considered during wave reconstruction and (b) the space-time diagram formulated according to the extreme components within these bounded limits.}
    \label{space_time_diagram}
\end{figure}

Defining the measuring point of the wave elevation as $x_{\scriptscriptstyle M} = 0$ and considering Eq. \ref{WaveEquation}, the recorded wave height can be denoted as $\zeta_{M} = \zeta(0,\left[ t_{0},t_{0}+\Delta t, \dots, T_M\right])$ where $\Delta t$ is a discrete sampling time interval. The frequency spectrum of $\zeta_{M}$ is obtained as:
\begin{equation}
    \mathcal{F}_{n} = \sum_{j=0}^{J-1} \zeta_{M}(j) \exp^{-i(2\pi \frac{jn}{J})}
\end{equation}
where $J = T_{M}/\Delta t$ is the number of wave-height samples in the wave-height record; an example of this is shown in Fig. \ref{space_time_diagram}(a). These Fourier coefficients produce the spectral parameters:
\begin{equation}
    A_{n} = |\mathcal{F}_{n}| \quad \quad \epsilon_{n} = \angle \mathcal{F}_{n} \quad \quad k_{n} = \omega_{n}^{2}/g \nonumber
\end{equation}
\begin{equation}
    \omega_{n} \in \left[ \frac{2\pi n}{J\Delta t}, \frac{2\pi (n+1)}{J\Delta t}, \dots,  \frac{2\pi}{J\Delta t}(J-1) \right] \nonumber
\end{equation}
which can be directly substituted into Eq. \ref{WaveEquation} to produce a predicted sea surface elevation $\Tilde{\zeta}(x_{\scriptscriptstyle V}, t)$, where $x_{\scriptscriptstyle V}$ is the location of the vehicle in the x-plane. For shallow water waves, the limit of the dispersion relation takes the form $\lim_{d \to \infty} \kappa^{2} \to \omega^2/gd$, resulting in a linear phase filter where $\kappa_{n}x_{\scriptscriptstyle V}$ is considered a phase-shifting filter parameter for propagating the wave through space \cite{Belmont2006}.


The key component which dictates what is referred to as the \emph{predictable region} (the length of time for which a valid wave prediction can be obtained) is the wave celerity, and in particular the celerity of the spectral components with the lowest and highest frequency; this is shown in Fig. \ref{space_time_diagram}(b). Denoting the celerity of these components $c_{min}$ and $c_{max}$ respectively, the maximum valid predictable region is a subset of the time-frame $t_{s} \to t_{f}$, where:
\begin{equation} \label{dswp_time_period}
    t_{s} = \frac{x_{P}}{c_{min}}, \quad \quad t_{f} = \frac{x_{P}}{c_{max}} + T_{M}
\end{equation}
and the definition of celerity varies relative to the seabed depth to wavelength ratio, \cite{ReeveBook}. Importantly, the variables $T_M$ and $t_f$ dictate a cut-off temporal threshold for the time scale of the receding horizon within the NMPC. The spectrum bounding technique can be seen in Fig. \ref{space_time_diagram}(a), demonstrating how this effects the maximum predictable region in Fig. \ref{space_time_diagram}(b).

In conjunction with estimating the wave elevation, the spectral information can provide an estimate of the particle velocity vector $\Tilde{\boldsymbol{\nu}}_{p} = \begin{bmatrix} \Tilde{\nu}_{p,x}, \Tilde{\nu}_{p,z} \end{bmatrix}$ for direct substitution into Eq. \ref{wave_forces}, yielding a prediction of the wave-induced disturbances $\Tilde{\boldsymbol{\tau}}_{E}$.


\subsection{NMPC Formulation}

The nonlinear dynamics in Eq. \ref{dynamic_equation} can be described  by the function:
\begin{align} \label{nmpc_ode}
    \mathbf{\dot{x}} & = f(\mathbf{x},\boldsymbol{\mu)}  \\
     & = \begin{bmatrix}
    \mathbf{J}(\boldsymbol{\eta})\boldsymbol{\nu} \\
    \mathbf{M}^{-1}(\mathbf{B}_{\boldsymbol{\mu}}\boldsymbol{\mu} + \boldsymbol{\tau}_{E} - \mathbf{C}(\boldsymbol{\nu})\boldsymbol{\nu} - \mathbf{D}(\boldsymbol{\nu})\boldsymbol{\nu} - \mathbf{g}(\boldsymbol{\eta})) 
    \end{bmatrix} \nonumber
\end{align}
where $\mathbf{x} = [ \boldsymbol{\eta}, \boldsymbol{\nu} ]^{T}$ and $\boldsymbol{\mu}$ represent the system state and generalised control input. Eq. \ref{nmpc_ode} is time-discretized and solved via Fourth Order Runge-Kutta integration:
\begin{equation}
    \mathbf{x}_{k+1} = F(\mathbf{x}_{k},\boldsymbol{\mu_{k}})
\end{equation}
where $k$ represents the current time-step and steps are discretized by $\Delta t$. 

From this, the NMPC can be formulated for reference tracking using a quadratic cost function. Inclusion of a terminal cost and constraint assists in achieving recursive feasibility \cite{Mayne2000}, producing:
\begin{align}
\mathcal{J} = & (\mathbf{x}_{k+N_{c}} - \mathbf{x}_{r})^{T}\mathbf{P}_{\mathbf{x}}(\mathbf{x}_{k+N_{c}} - \mathbf{x}_{r}) +\\
 & \sum_{k}^{k+N_{c}-1} (\mathbf{x}_{k} - \mathbf{x}_{r})^{T}\mathbf{Q}_{\mathbf{x}}(\mathbf{x}_{k} - \mathbf{x}_{r}) + \nonumber \\ 
& \sum_{k}^{k+N_{c}-1} \boldsymbol{\mu}_{k}^T \mathbf{R}_{\mu} \boldsymbol{\mu}_{k} + \Delta\boldsymbol{\mu}_{k}^T \mathbf{R}_{\mu} \Delta\boldsymbol{\mu}_{k} \nonumber 
\label{nmpc_cost_function}
\end{align}
where $\mathbf{x}_{r}$ refers to the reference state which is time varying in these instances. Here, $\mathbf{P}_{\mathbf{x}} \in \mathbb{R}^{3\times 3}$, $\mathbf{Q}_{\mathbf{x}}\in \mathbb{R}^{3\times 3}$ and $\mathbf{R}_{\mu}\in \mathbb{R}^{3\times 3}$ are weighting matrices on the terminal state, intermediate state and control. For the results provided in Section \ref{results}, $N_{c}=20\Delta t$, $\mathbf{P}_{\mathbf{x}} = \mathbf{Q}_{\mathbf{x}} = \text{diag}(250,250,250)$ and $\mathbf{R}_{\mu} = \text{diag}(1,1,1)$, where the weighting matrices were tuned heuristically using Bryson's rule \cite{Bryson1975} as a first approximation. By adjusting the magnitude of these weighting matrices, higher priority can be placed on either state regulation or energy conservation based on the demands of the mission, varying the controller performance consequently.

From this, the optimal control problem can be formulated as:
\begin{mini}|l|
	  {}{\quad \mathcal{J}(\mathbf{x}_{k},\boldsymbol{\mu}_{k})}{}{}
	  \addConstraint{\quad \mathbf{x}_{k+1}}{= F(\mathbf{x}_{k},\boldsymbol{\mu}_k)}
	  \addConstraint{\quad \mathbf{x}_{0}}{= \mathbf{x}(t_{0}) }
        \addConstraint{\quad \mathbf{x}_{k}}{\in \boldsymbol{\mathcal{X}}}
        \addConstraint{\quad \boldsymbol{\mu}_{k}}{\in \boldsymbol{\mathcal{U}} }
	  \addConstraint{\quad \mathbf{x}_{k+N}}{\in \boldsymbol{\mathcal{X}}_{f} }
\end{mini}
\noindent where $\boldsymbol{\mathcal{X}}$, $\boldsymbol{\mathcal{U}}$ and $\boldsymbol{\mathcal{X}}_{f}$ are the feasible sets for the stage and terminal components respectively. The terminal set was defined as a constant radius region surrounding the trajectory set-point at time-step $k$, rather than a "zero error" terminal constraint which is less practical and more restrictive, particularly when considering disturbances as in this work. Implementation of the NMPC scheme was performed using the CasADi MATLAB library in conjunction with the IPOPT solver, using a direct multiple shooting method which is known to improve convergence rates by solving the optimisation problem in smaller intervals in contrast to a single shooting method \cite{Andersson2019}. 

\section{Results}
\label{results}

\subsection{Scenario Configuration}

The environment in our analysis is configured to be representative of typical conditions the ROV may encounter during an inspection task for a shallow water offshore structure, for example a wind turbine monopile or jacket. As alluded to earlier, the concern here is the upper portion of the water column where wave effects have a greater influence over the vehicle behaviour, thus reliable and accurate control is required to mitigate perturbations. The method presented in Section \ref{wave_predictor} requires temporal data of the sea surface elevation, therefore spectral data collected by a real-world wave buoy was exploited in this analysis. More specifically, a buoy located near an active offshore wind farm was selected where the seabed depth is $d = 54$m. The vehicle was situated at an operational depth of $z=5$m and an initial distance from the buoy of $x_{\scriptscriptstyle P} = 50$m. Three separate wave conditions were simulated, each with different peak spectral periods and similar wave heights - the specific spectral characteristics are given in Table \ref{spectral_parameters}. These cases were chosen to demonstrate operation under a variety of real-world conditions with substantial wave height, with an extensive systematic behavioural study planned for future work to characterise the vehicle response over specified intervals. Similarly, the vehicle characteristics are provided in Table \ref{BlueROV2_Parameters} where $X_{\dot{q}}$ and $M_{\dot{u}}$ were evaluated numerically using the potential flow solver WAMIT\copyright.

\begin{table}[t!] 
\renewcommand{\arraystretch}{1.3}
\caption{Statistical parameters for the 3 wave spectra considered in this work, varying in peak spectral period and significant wave height.}.
\label{spectral_parameters}
\centering
\begin{tabular}{|c|c|c|} 
\hline
Case Reference  & Peak Period (s) & Significant Wave Height (m) \\
\hline
W1 & 7.1 & 2.78\\
W2 & 9.5 & 3.47 \\
W3 & 11.1 & 3.24 \\
\hline
\end{tabular}
\end{table}
\begin{table}[t!]
\renewcommand{\arraystretch}{1.3}
\caption{BlueROV2 Heavy dimensions and hydrodynamic parameters utilised in the simulations; data based on \cite{Benzon2021, brov2, Wu2018}.}
\label{BlueROV2_Parameters}
\centering
\begin{tabular}{|c|c|c|} 
\hline
Parameter & Nomenclature & Value \\
\hline
Weight & $W$ & 112.8 N\\
Buoyancy & $B$ & 114.8 N\\
Rotational Inertia, $y$ & $I_{y}$ & 0.253 kgm$^{2}$ \\
Added Inertia Coeff. & $X_{\dot{u}}$, $Z_{\dot{w}}$ & 6.36, 18.68 kg\\
" & $M_{\dot{q}}$ & 0.135 kgm$^{2}$\\
" & $X_{\dot{q}}$, $M_{\dot{u}}$ & 0.67 kgm \\
Linear Drag Coeff. & $X_{u}$, $Z_{w}$ & 13.7, 33 kg/s\\
" & $M_{q}$ & 0.80 kgm$^{2}$/s\\
Quadratic Drag Coeff. & $X_{u|u|}$, $Z_{w|w|}$ & 141, 190  Ns$^{2}$/m$^{2}$\\
" & $M_{q|q|}$ & 0.47 Nms$^{2}$ \\
Centre of Buoyancy & $r_{B}$ & [0, 0, 0.028] m \\
Maximum Thrust & $T_{max}$ & 35 N \\
Thruster Offset & $\alpha$ & 45$^{o}$ \\
\hline
\end{tabular}
\end{table}

The proposed framework was tested for a specified mission of following a square trajectory while subject to disturbances from each of wave field reported in Table \ref{spectral_parameters}; this could relate to a typical inspection task over a structural area of interest. The start and end point of the tracking mission was specified at $(50, -8)$m with respect to the prediction point at $(0, 0)$m (the free surface). The mission was set to translate forward $5$m longitudinally, ascend $5$m, translate backward $5$m longitudinally and finally descend $5$m back to the original position, all whilst minimising pitch displacement. Each simulation was undertaken over a 600s temporal segment with a resolution of $\Delta t = 0.1$s; the initial 300s were used to accumulate a temporal record of the wave elevation, after which continuous wave predictions are produced for the final 300s to undertake the trajectory tracking mission. The reference state of the vehicle was specified as a vector of uniformly spaced points between each corner of the trajectory and was defined prior to beginning the mission. 



\begin{figure}[t]
    \centering
    \includegraphics[width=0.48\textwidth]{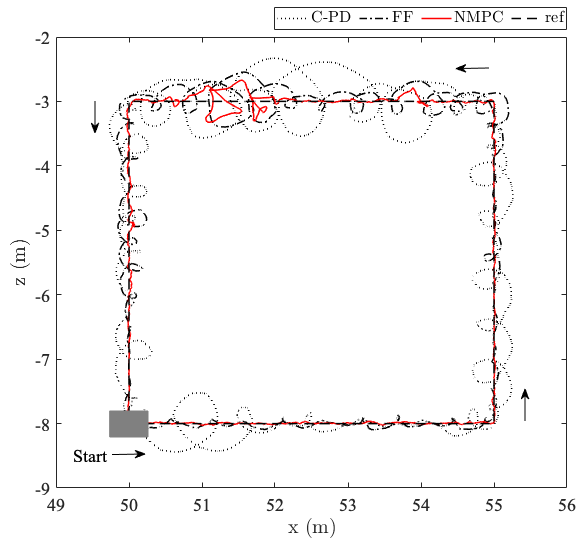}
    \caption{Trajectory tracking spatial performance of each controller for case W1, showing the NMPC successfully suppressing the cyclical wave-induced loads for the majority of the mission, even at low operational depth.  }
    \label{traj_spacial}
\end{figure}

\begin{figure}[t]
    \centering
    \includegraphics[width=0.48\textwidth]{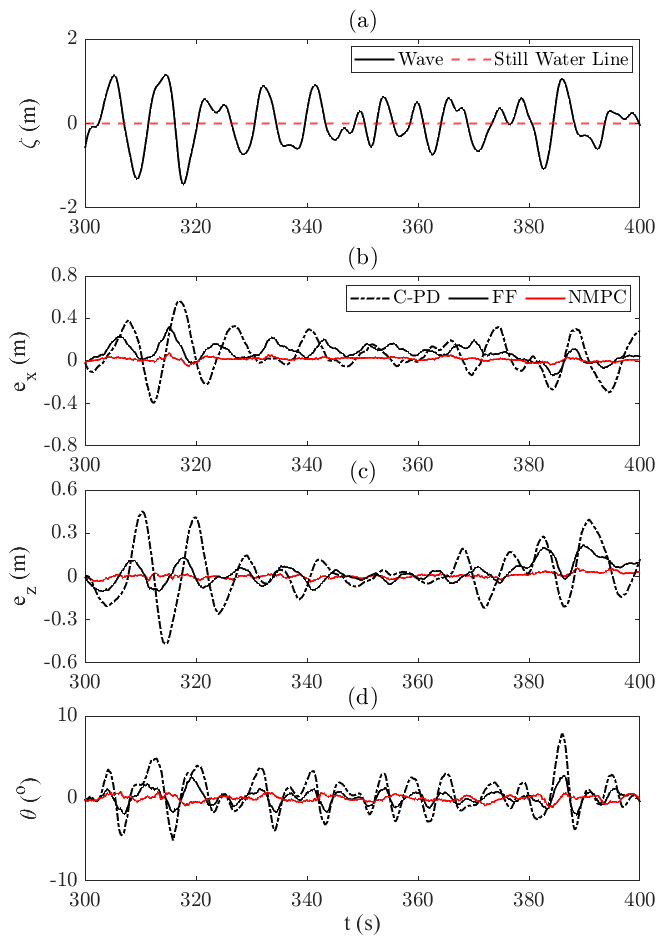}
    \caption{Temporal segment of the trajectory tracking mission for the (a) wave in case W2, displaying the positional error relative to the trajectory in the (b) surge, (c) heave and (d) pitch. }
    \label{traj_time_history}
\end{figure}

To present a comparable study, a cascaded position-velocity proportional-derivative controller (referred to as C-PD from henceforth) and a C-PD with feed-forward (FF) controller (referred to as FF from henceforth) were also implemented and analysed. The former is to provide a basic reference against a standard controller, with the latter incorporating the prediction algorithm to calculate disturbance compensating control actions without the optimisation loop on the control trajectory. The generalised control laws for the C-PD controller and FF controller are therefore defined as:
\begin{equation} \label{PIDcontrol}
    \boldsymbol{\mu}_{PD} = \mathbf{K}_{p,v} \lbrace \boldsymbol{\nu} - \left( \mathbf{K}_{p}\mathbf{e} + \mathbf{K}_{d}\mathbf{\dot{e}} \right) \rbrace
\end{equation}
\begin{equation}
    \boldsymbol{\mu}_{FF} = \boldsymbol{\mu}_{PD} + \boldsymbol{\tau}_{max}^{-1} \odot \left( \mathbf{M_{A}\dot{\boldsymbol{\nu}_{p}}} + \mathbf{D}(\boldsymbol{\nu}_{p})\boldsymbol{\nu}_{p} \right)
\end{equation}
where $\mathbf{e} = \mathbf{x}_{r} - \mathbf{\hat{x}}_{k}$ is the proportional error between the set-point and estimated state obtained via the EKF.

\subsection{Trajectory Tracking Performance}

An example relating to case W2 is depicted in Fig. \ref{traj_spacial}, showing the direction of motion and the performance of the three different control strategies in tracking the reference. The circular trajectories identify the wave-induced fluid motions the vehicle is subject to, which are naturally elliptical in shallow water and decay exponentially with depth \cite{McCormickBook}. This explains why the displacements at $3$m depth are much more pronounced than elsewhere during the mission. Qualitatively it is apparent from Fig. \ref{traj_spacial} that the NMPC performs best, with the vehicle very rarely diverging from the reference. 
The temporal evolution for case W2 is shown in Fig. \ref{traj_time_history}, comparing the error in the three DoFs under the wave profile reported in Fig. \ref{traj_time_history}(a).

\begin{figure}[t]
    \centering
    \includegraphics[width=0.48\textwidth]{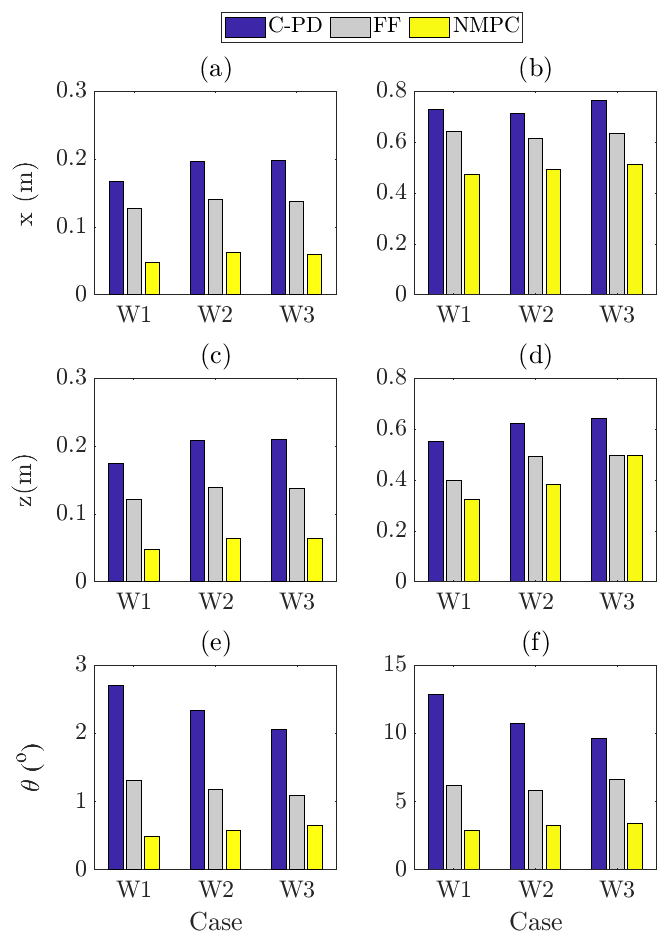}
    \caption{ (a)(c)(e) RMSE and (b)(d)(f) maximum error for each DoF and sea state considered for the trajectory tracking mission.}
    \label{traj_rmse}
\end{figure}

Quantitative measures of performance were also considered in the form of RMSE and maximum error exhibited throughout the trajectory tracking mission, given in Fig. \ref{traj_rmse}. At minimum, there was a reduction in RMSE between the NMPC and FF strategies of $\approx 47\%$, $\approx 49\%$ and $\approx 27\%$ in the surge, heave and pitch respectively and across all cases; the additional preview information along the NMPC prediction horizon facilitates this significant improvement. The length of the prediction horizon within the NMPC is highly dependent on the length of disturbance preview available, thus the performance is also dependent on this factor. Naturally this therefore affects performance, but as the proposed strategy optimises control actions in response to disturbance estimations, even shorter horizons would offer improved performance over classical feedback methods. With regards to maximum error, the only exception to NMPC superior performance was related to the heave DoF of case W3, with a marginal rise in error of $\approx 5\%$. For all other instances, the maximum error was reduced on average by $12\%$ with respect to positional tracking and $46\%$ with respect to attitude regulation. The largest increases in performance relate to the pitch, with a reduction of up to $\approx 58\%$ and $\approx 49\%$ in RMSE and maximum error, both relating to case W3. In general, performance was fairly consistent irrespective of spectral period, showing a good level of robustness to different sea states.


\begin{figure}[t]
    \centering
    \includegraphics[width=0.48\textwidth]{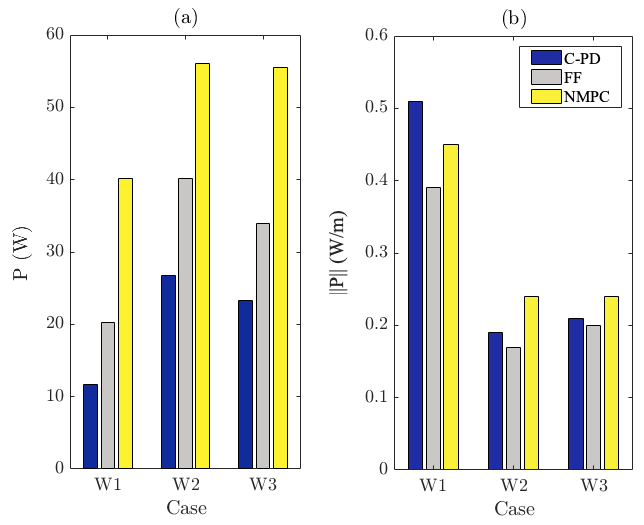}
    \caption{ The (a) power consumed during the trajectory tracking mission and (b) the power consumption relative to normalised RMSE error, where the normalisation is $1/$RMSE. }
    \label{traj_power}
\end{figure}

Across all cases considered in this work, the power consumed naturally increases when deploying the predictive control methods, Fig. \ref{traj_power}(a). A comparison of relative power consumption to normalised RMSE is displayed in Fig. \ref{traj_power}(b), which provides a metric to rigorously compare the improved control performance against added power expenditure. This relative ratio shows similar behaviour across all wave cases, fluctuating between $0.4-0.5$W/m for case W1 and $\approx 0.2$ W/m for the other two cases. This supports the statement that the additional power consumed by the predictive strategies is utilised effectively, with longer wavelengths exhibiting comparable behaviour. This is interesting, as it could indicate a potential inflection point of spectral period where the performance increase to consumed power relationship plateaus. For calmer conditions, the FF method may be more applicable to sufficiently track the trajectory well with less computational overhead; relative to standard feedback control, the FF scheme still offers a reasonable improvement in performance, which could be useful for missions where power conservation is critical.




\section{Conclusions}

To tackle the problem of trajectory tracking for underwater vehicles in the presence of wave disturbances, this paper has proposed a model predictive control strategy which explicitly considers online forecasting of future wave perturbations. Incorporating these within the control optimisation improved RMSE (on average) by $51\%$ for positional tracking (surge/heave) and $44.5\%$ for attitude regulation when compared to an alternative feed-forward compensating method. The strategy was tested under three different scenarios, demonstrating effective performance independent of wave conditions; significant wave heights were of magnitude several times larger than the vehicle length (at minimum 6$\times$) in all cases, proving the controller's ability to mitigate large wave loads. A power analysis was also performed, showing marginal difference in normalised power expenditure relative to displacement. This paper has provided preliminary evidence of the potential of the proposed approach based on Deterministic Sea Wave Prediction, pointing at a viable, realistic solution to the problem of robust operation in hazardous wave climates.







\end{document}